\documentclass{article} 
\usepackage{iclr2017_conference,times}
\usepackage{hyperref}
\usepackage{url}
\usepackage{amsmath}
\usepackage{graphicx}
\usepackage{subfigure}
\usepackage{lipsum}
\usepackage{amssymb}
\usepackage[titletoc,title]{appendix}

\DeclareMathOperator{\sinc}{sinc}
\usepackage{caption}

\title{\textbf{}}
\author{Rajesh P. N. Rao}
\date{\vspace{-7ex}}
\newcommand{\nop}[1]{}

\title{Transformational Sparse Coding}

\author{Dimitrios C. Gklezakos \& Rajesh P. N. Rao  \\
Department of Computer Science\\
and Center for Sensorimotor Neural Engineering\\
University of Washington\\
Seattle, WA 98105, USA \\
\texttt{\{gklezd,rao\}@cs.washington.edu} \\
}

\begin{document}

\maketitle

\begin{abstract}
A fundamental problem faced by object recognition systems is that
objects and their features can appear in different locations, scales
and orientations. Current deep learning methods attempt to achieve
invariance to local translations via pooling, discarding the locations
of features in the process.  Other approaches explicitly learn
transformed versions of the same feature, leading to representations
that quickly explode in size. Instead of discarding the rich and
useful information about feature transformations to achieve
invariance, we argue that models should learn object features
conjointly with their transformations to achieve equivariance.  We
propose a new model of unsupervised learning based on sparse coding
that can learn object features jointly with their affine
transformations directly from images. Results based on learning from
natural images indicate that our approach
matches the reconstruction quality of traditional sparse coding but
with significantly fewer degrees of freedom while simultaneously
learning transformations from data. These results open the door to
scaling up unsupervised learning to allow deep feature+transformation
learning in a manner consistent with the ventral+dorsal stream
architecture of the primate visual cortex.
\end{abstract}

\section{Introduction}
A challenging problem in computer vision is the reliable recognition
of objects under a wide range of transformations. Approaches such as
deep learning that have achieved success in recent years usually
require large amounts of labeled data, whereas the human brain has
evolved to solve the problem using an almost unsupervised approach to
learning object representations. During early development, the brain
builds an internal representation of objects from unlabeled images
that can be used in a wide range of tasks.  

Much of the complexity in learning efficient and general-purpose
representations comes from the fact that objects can appear in
different poses, at different scales, locations, orientations and
lighting conditions. Models have to account for these transformed
versions of objects and their features. Current successful approaches
to recognition use pooling to allow limited invariance to
two-dimensional translations (\cite{ranzato-cvpr-07}). At the same time
pooling discards information about the location of the detected
features. This can be problematic because scaling to large numbers of
objects requires modeling objects in terms of parts and their relative
pose, requiring the pose information to be retained.

 Previous unsupervised learning techniques such as sparse coding
 (\cite{olshausen1997sparse}) can learn features similar to the ones in
 the visual cortex but these models have to explicitly learn large
 numbers of transformed versions of the same feature and as such,
 quickly succumb to combinatorial explosion, preventing hierarchical
 learning. Other approaches focus on computing invariant object
 signatures (\cite{DBLP:journals/corr/AnselmiLRMTP13,Anselmi:2016:ULI:2951995.2952046}), but are
 completely oblivious to pose information. 

Ideally, we want a model that allows object
features and their relative transformations to be {\em simultaneously
  learned}, endowing itself with a combinatorial explanatory
capacity by being able to apply learned object features with
object-specific transformations across large numbers of objects. The
goal of modeling transformations in images is two-fold: (a) to
facilitate the learning of pose-invariant sparse feature
representations, and (b) to allow the use of pose information of
object features in object representation and recognition.

We propose a new model of sparse coding called {\em transformational
  sparse coding} that exploits a tree structure to account for large
affine transformations. We apply our model to natural images.
We show that our model
can extract pose information from the data while matching the
reconstruction quality of traditional sparse coding with significantly
fewer degrees of freedom.  Our approach to unsupervised learning is
consistent with the concept of ``capsules'' first introduced by \cite{Hinton:2011:TA:2029556.2029562}, and more
generally, with the dorsal-ventral (features+transformations)
architecture observed in the primate visual cortex.

\section{Transformational Sparse Coding}\label{model}
\subsection{Transformation Model}
Sparse coding (\cite{olshausen1997sparse}) models each image $I$ as a sparse combination of features:
$$
I \simeq Fw \:\:\: \text{s.t. w is sparse}
$$
Sparsity is usually enforced by the appropriate penalty. A typical choice is $S_1(w) = \|w\|_1$.
We can enhance sparse coding with affine transformations by transforming features before combining them. The vectorized input image $I$
is then modeled as:
$$
I = \sum_{k=1}^{K} w_{k} T(x_{k})F_k
$$
where $w_k,F_k$ denote the $k$-th weight specific to the image and the $k$-th feature respectively and $T(x_{k})$
is a feature and image specific transformation.

In modeling image transformations we follow the approach of \cite{Rao:1999:LLG:340534.340807} and \cite{citeulike:2714601}.
We consider the $2$D general affine transformations. These include rigid motions such as vertical and horizontal translations and rotations,
as well as scaling, parallel hyperbolic deformations along the $X/Y$ axis and hyperbolic deformations along the diagonals. A discussion on why these
are good candidates for inclusion in a model of visual perception can be found in \cite{Dodwell1983}. Figure \ref{fig:G_effects} in Appendix \ref{app:Gs}
shows the effects of each transformation.

\begin{figure}[ht!]
\begin{center}
    \subfigure[]{
		\includegraphics[height=0.26\textwidth,width=0.3\textwidth]{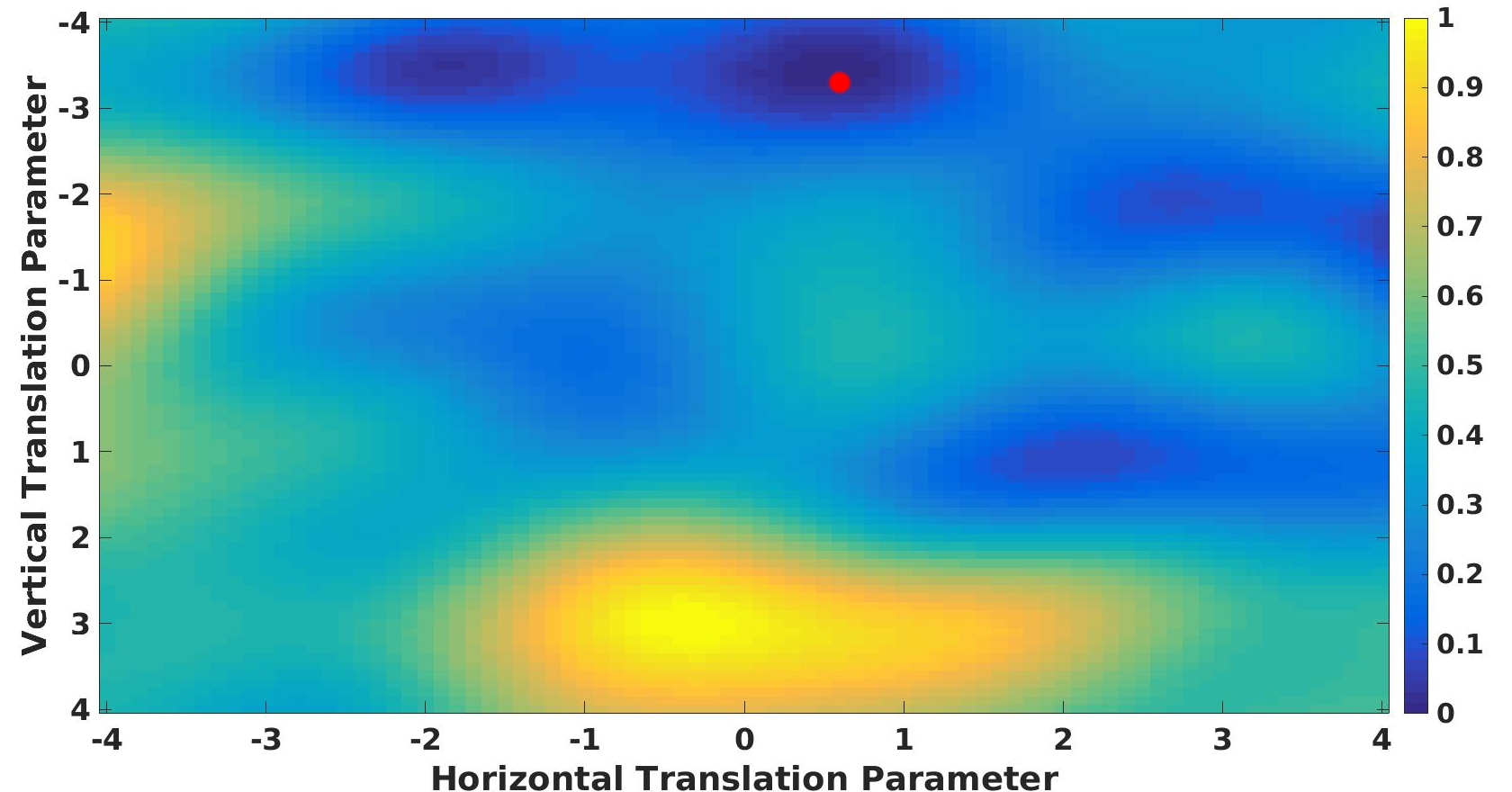}
		\label{fig:xys}
    }
    \subfigure[]{
		\includegraphics[height=0.26\textwidth,width=0.3\textwidth]{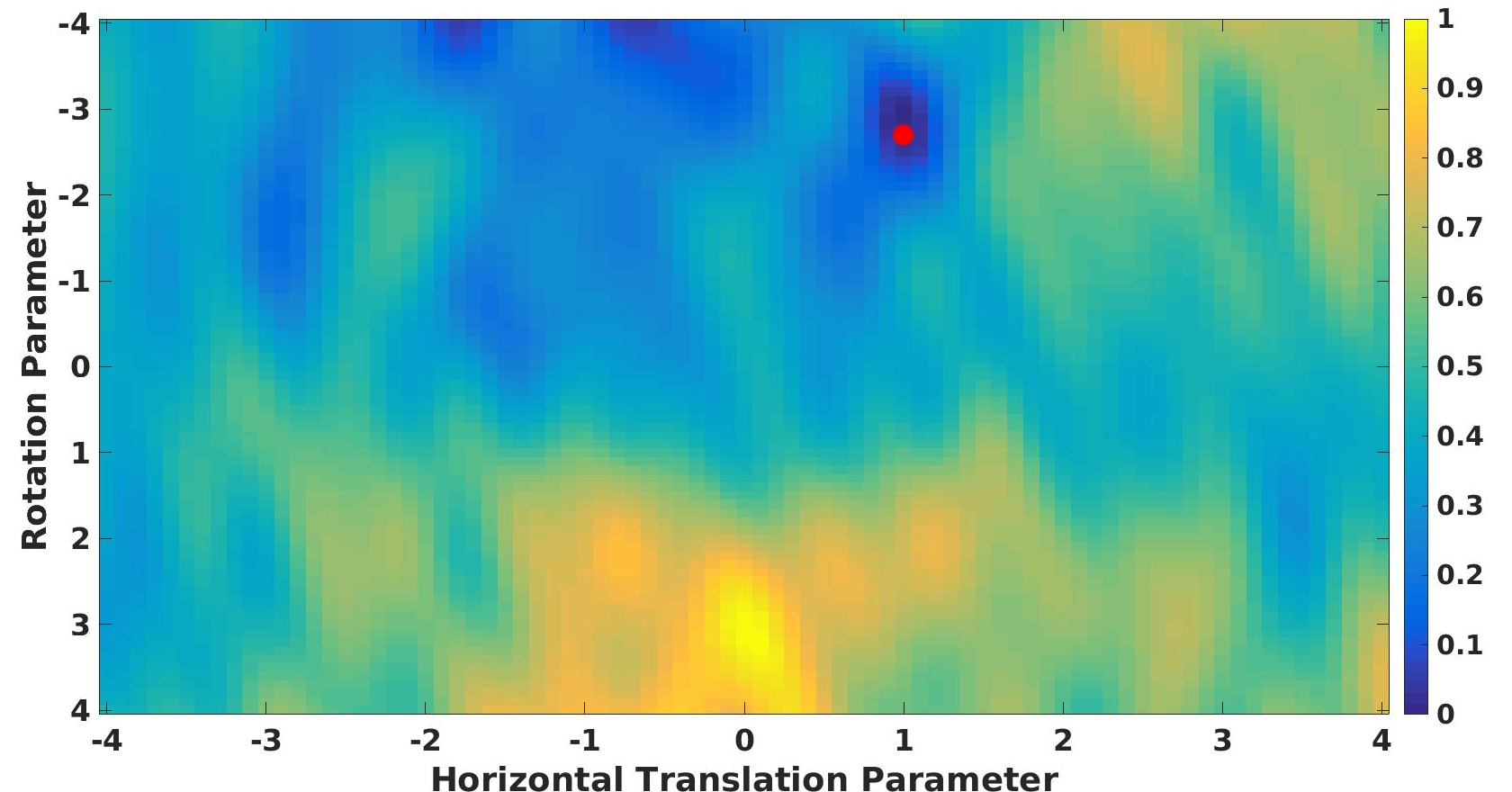}
		\label{fig:xrs}
    }
    \subfigure[]{
		\includegraphics[height=0.26\textwidth,width=0.3\textwidth]{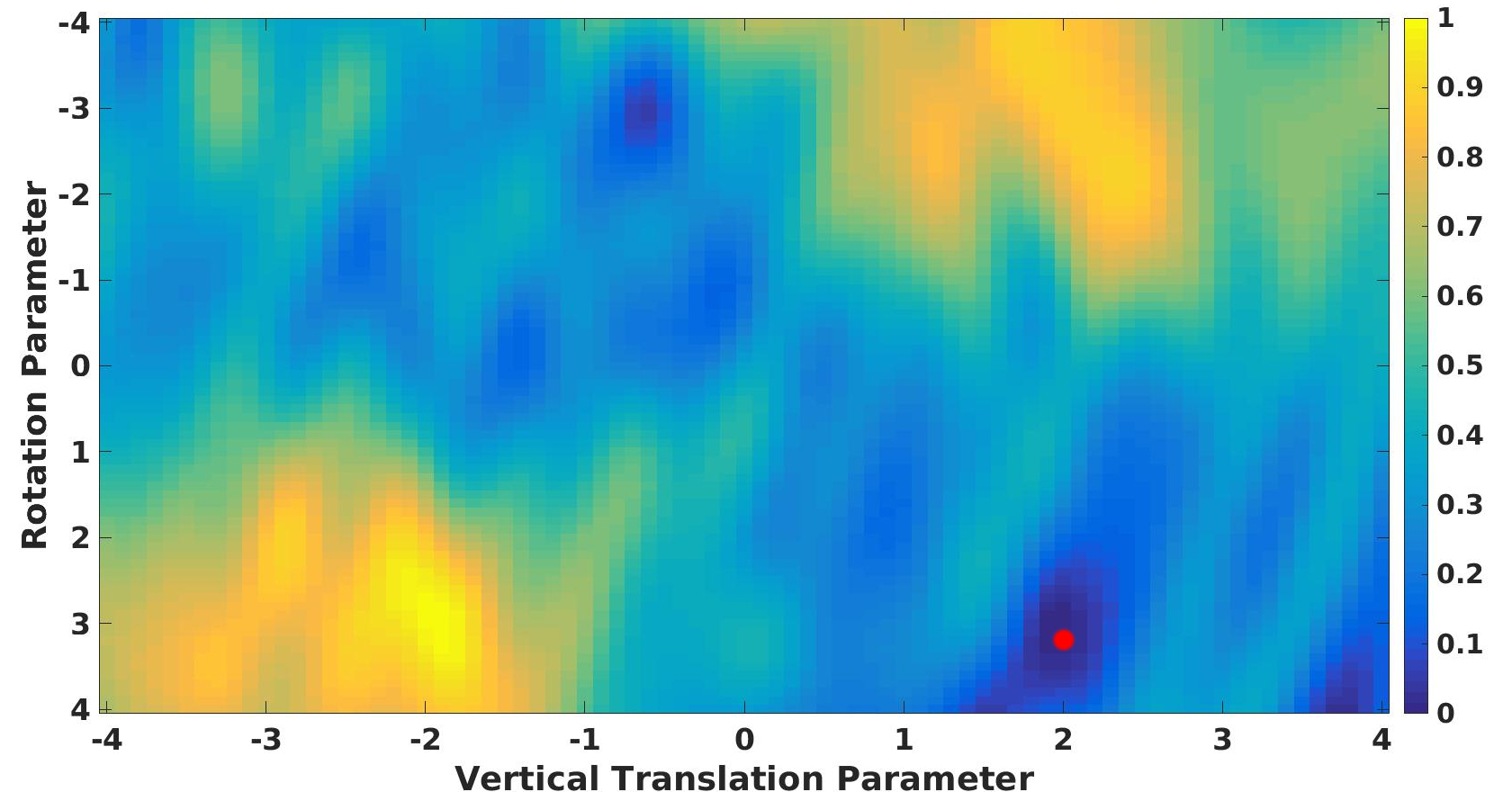}
		\label{fig:yrs}
    }\\
    \subfigure[]{
		\includegraphics[height=0.26\textwidth,width=0.3\textwidth]{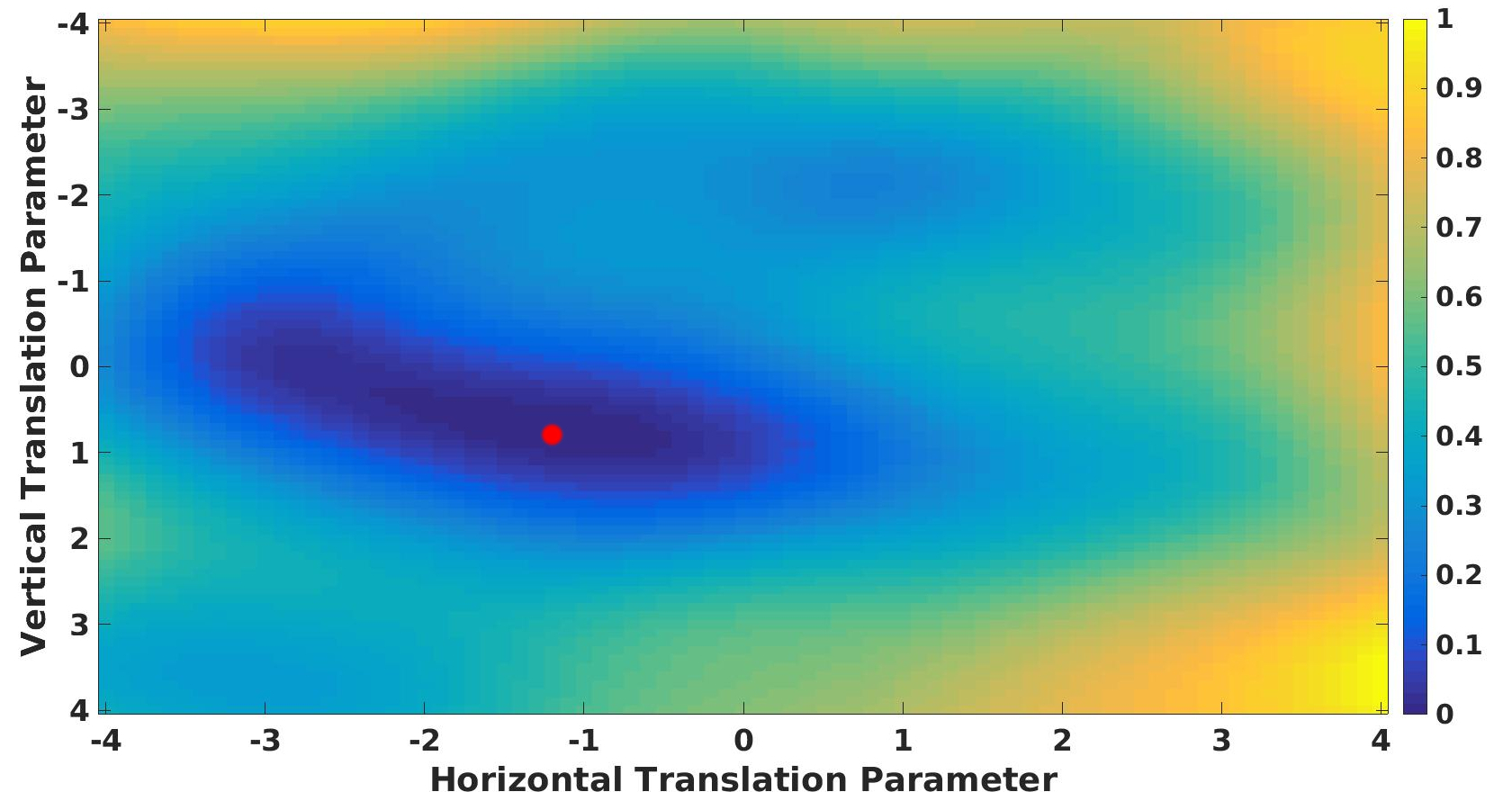}
		\label{fig:xyb}
    }
    \subfigure[]{
		\includegraphics[height=0.26\textwidth,width=0.3\textwidth]{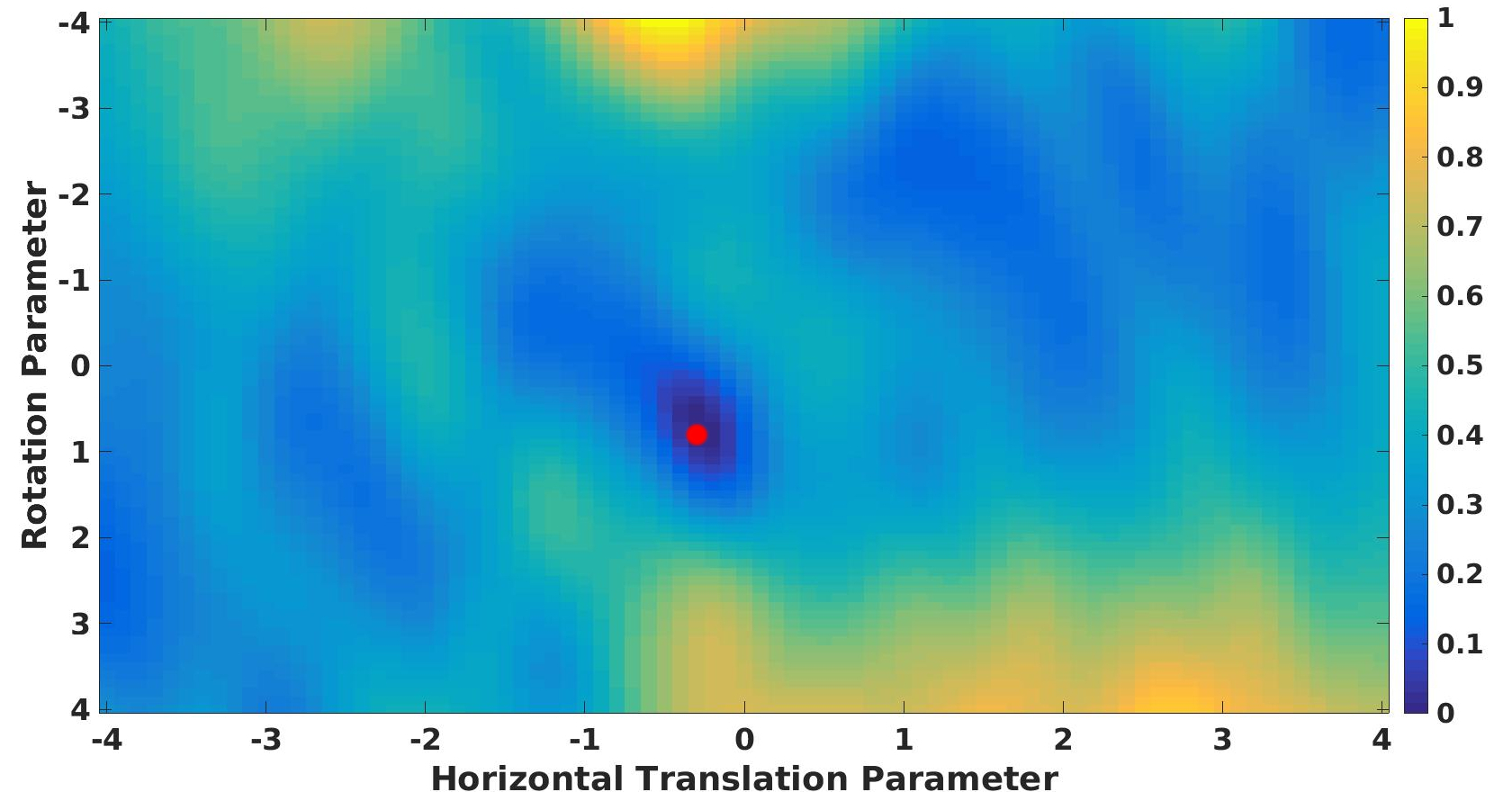}
		\label{fig:xrb}
    }
    \subfigure[]{
		\includegraphics[height=0.26\textwidth,width=0.3\textwidth]{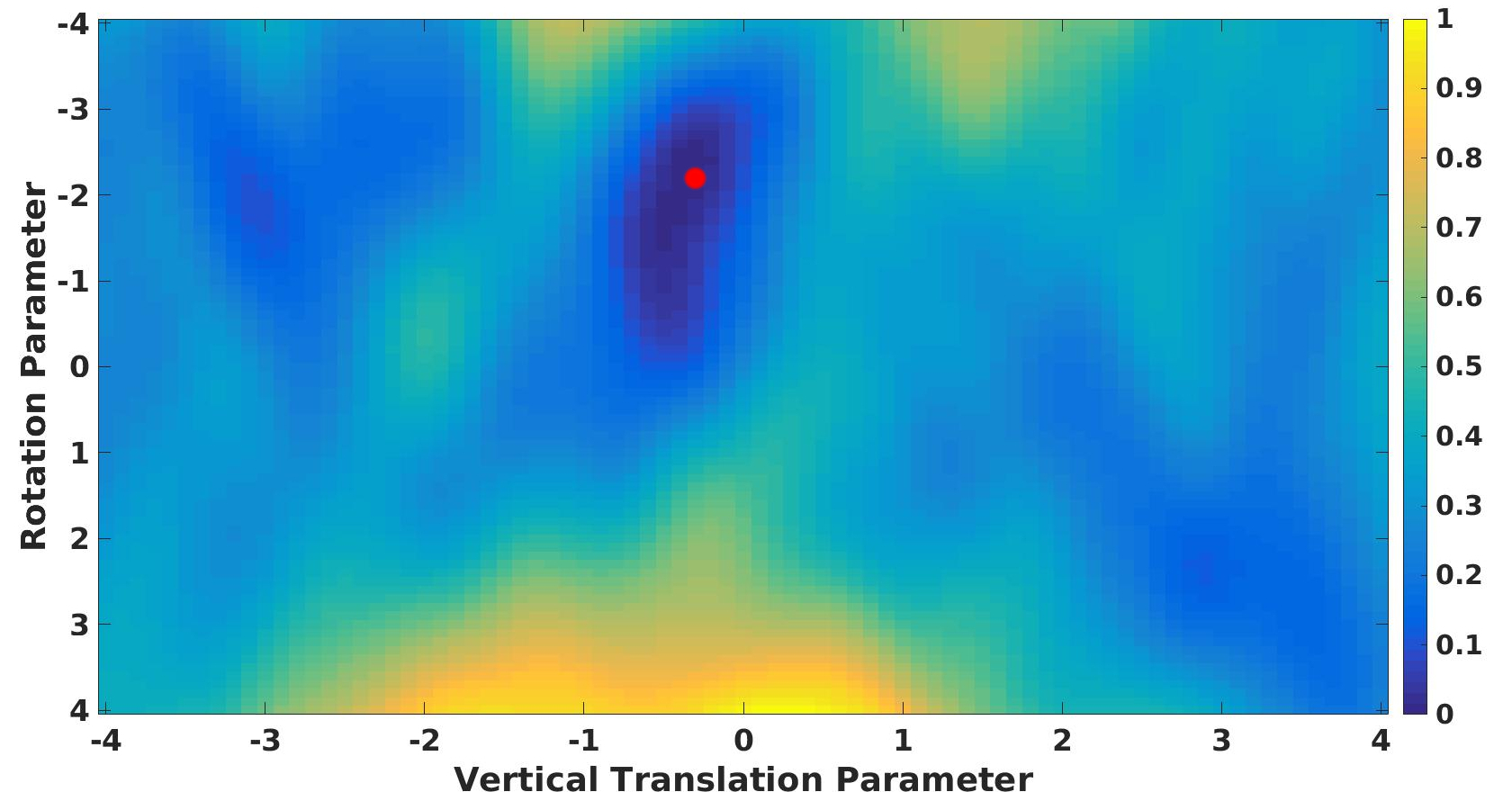}
		\label{fig:yrb}
    }
\end{center}
\caption{Normalized reconstruction error for individual vs. batch $8\times 8$ natural image patches. (a),(b),(c) show the surface of the reconstruction error
for horizontal and vertical translations, horizontal translations and rotation, vertical translations and rotations for an individual data point and feature. (d),(e),(f) show the same,
averaged over a batch of $2000$ data points. The error is normalized between $0$ and $1$ for comparison. The global minimum in the range is marked in red. In the batch case, averaging makes the error surface smoother and learning easier.}
\label{fig:general_loss}
\end{figure}

Any subset of these transformations forms a Lie group with the corresponding number of dimensions ($6$ for the full set).
Any transformation in this group can be expressed as the matrix exponential of a weighted combination of matrices (the group generators) that describe
the behaviour of infinitesimal transformations around the identity:
$$
T(x) = e^{\sum_j x_j G_j}
$$
For images of $M$ pixels, $T(x)$ is a matrix of size $M\times M$. Note that the generator matrices and the features used are common
across all images. The feature weights and transformation parameters can be inferred (and the features learned) by gradient descent on the
regularized MSE objective:
$$
L(w,x,F) = \frac{1}{N}\sum_{i=1}^{N}\left\Vert I_i - \sum_{k=1}^{K}w_{ik}T(x_{ik}) F_k \right\Vert_2^2 + \lambda_w S_1(w) + \lambda_F \|F\|_2^2
$$
Although this model ties sparse coding with transformations elegantly, learning large transformations with it is intractable.
The error surface of the loss function is highly non-convex with many shallow local minima. Figures \ref{fig:xys},\ref{fig:xrs},\ref{fig:yrs} show the surface of $L$
as a function of horizontal and vertical translation, horizontal translation and rotation and vertical translation and rotation parameters. The model tends to
settle for small transformations around the identity. Due to the size of the parameters that we need to maintain, a random restart approach would be infeasible.

\subsection{Transformation Trees}

\begin{figure}[ht!]
\begin{center}
    \subfigure[]{
		
	\quad
	\quad
	\quad
	\includegraphics[height=0.3\textwidth,width=0.4\textwidth]{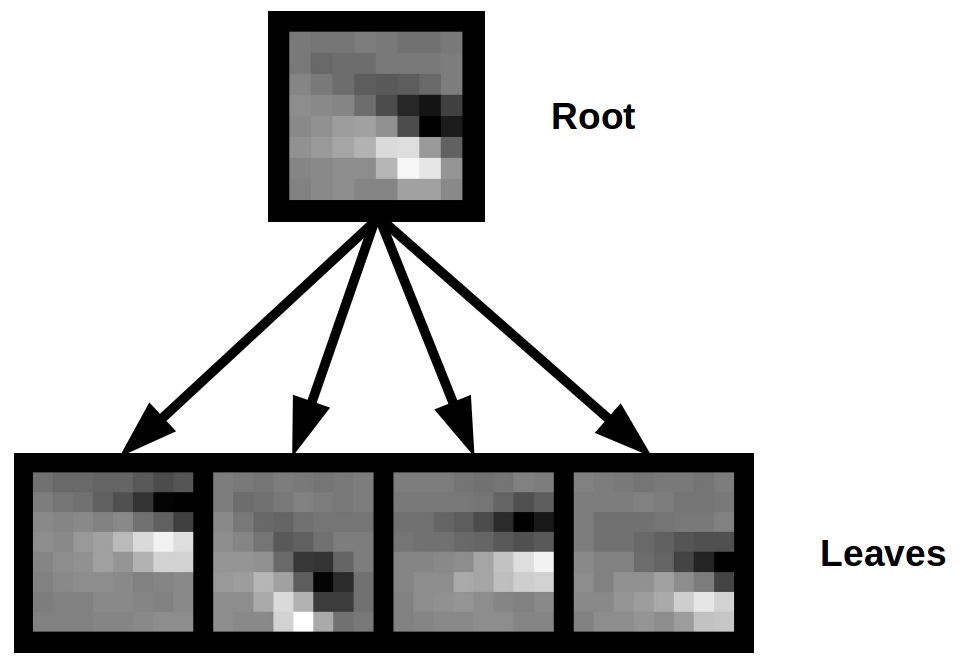}
	\label{fig:tree_example_sub}
    }
\end{center}
\caption{Example of a tree learned from natural image patches. The leaves correspond to rigid transformations of the root.}
\label{fig:tree_example}
\end{figure}

We introduce {\em Transformational Sparse Coding Trees} to circumvent
this problem using hierarchies of transformed features. The main
idea is to gradually marginalize over an increasing range of
transformations. Each node in the tree represents a feature derived as
a transformed version of its parent, with the root being the template of the feature. The leaves
are equivalent to a set of sparse basis features and are combined to reconstruct the input as described above.
A version of the model using a forest of trees of depth one (flat trees), is given by:
$$
I \simeq \sum_{v=1}^{V} \sum_{b\sim ch(v)}w_{b}U_b
$$
where $U_b = T(x_{v\rightarrow b})F_v$ and $ch(v)$ the children of root $v$.
The feature $U_b$ is a leaf, derived from the root feature $F_v$ via the fixed (across all data-points) transformation $T(x_{v\rightarrow b})$.
Deeper trees can be built accordingly (Section \ref{sec:deeper_trees}). A small example of a tree learned from natural image
patches is shown in Figure \ref{fig:tree_example}.

There are multiple advantages to such a hierarchical organization of
sparse features. Some transformations are more common in data than
others. Each path in the tree corresponds to a transformation that is common across images. Such a path can be
viewed as a ``transformation feature'' learned from the data. Each
additional node in the tree ``costs'' a fixed set of new parameters
equal in size to the dimensions of the underlying Lie group (six in
our case). At the same time the node contributes a whole
new feature to the sparse code. Averaging over many data points, smoothens the
surface of the error function and makes larger transformations more accessible
to optimization. Figures \ref{fig:xyb},\ref{fig:xrb},\ref{fig:yrb} show the error surface
averaged over a batch of $2000$ patches.

For every leaf that is activated, the root template represents the
identity of the feature and the transformation associated with the
path to the root, the pose. In other words the tree is an equivariant
representation of the feature over the parameter region defined by the
set of paths to the leaves, very similar to the concept of a capsule
introduced by \cite{Hinton:2011:TA:2029556.2029562}.  In
fact, every increasing subtree corresponds to a capsule of increasing
size.

\subsection{Learning}

The reconstruction mean squared-error (MSE) for a forest of flat trees is given by:
$$
L_{MSE}(w,x,F) = \frac{1}{N}\sum_{i=1}^{N}\left\Vert I_i -  \sum_{v=1}^{V} \sum_{b\sim ch(v)}w_{ib}T(x_{v\rightarrow b})F_v \right\Vert_2^2
$$
Increasing the feature magnitudes and decreasing the weights will result in a decrease in loss. We constraint the root feature magnitudes to be
of unit $\ell_2$ norm. Consider different, transformed, versions of the same root template. For every such version there is a set of tree parameters that compensates
for the intrinsic transformation of the root and results in the same leaves. To make the solution unique we directly penalize the transformation parameter
magnitudes. Since scaling and parallel deformation can also change the magnitude of the filter, we penalize them more to keep features/leaves close to unit
norm. The full loss function of the model is:
$$
L(w,x,F) = L_{MSE}(w,x,F) + \lambda_w S_1(w) + \sum_{j=1}^{6} \lambda_j X_{[j]}
$$
s.t.
$$
\forall v , \|F_v\|_2 = 1
$$
where $X_{[j]}$ is the vector of the collective parameters for generator $G_j$.

\cite{NIPS2006_2979} use an alternating optimization approach to sparse coding. First the weights are inferred using the feature sign algorithm and then the features
are learned using a Lagrange dual approach. We use the same approach for the weights. Then we optimize the transformation
parameters using gradient descent. The root features can be optimized  using the analytical solution and projecting
to unit norm.

The matrix exponential gradient $\frac{\partial L}{\partial x}$ can be computed using the following formula (\cite{2001IJNME..52.1431O}):
$$
\frac{\partial e^{A(t)}}{\partial t} = \int_{0}^{1} e^{\alpha A(t)}\frac{\partial A(t)}{\partial t}e^{(1-\alpha)A(t)}d\alpha
$$
The formula can be interpreted as:
$$
E_{\alpha\sim U(0,1)}\left[ D(\alpha)\right]
$$
where $D(\alpha) = e^{\alpha A(t)}\frac{\partial A(t)}{\partial t}e^{(1-\alpha)A(t)}$.
For our experiments we approximated the gradient by drawing a few samples \footnote{In practice even a single sample works well. The computation over samples is easily parallelizable.} $\{\tilde{\alpha}_s\}_{s=1}^{S}$
and computing $E_{\tilde{\alpha}}\left[ D(\tilde{\alpha})\right]$. This can be regarded as a stochastic version of
the approach used by \cite{NIPS2009_3791}.

Some features might get initialized near shallow local optima (i.e. close to the borders or outside the receptive field). These features eventually become under-used by the model.
We periodically check for under-used features and re-initialize their transformation parameters \footnote{A feature is under-used when the total number of data-points using it in a batch
drops close to zero.}. For re-initialization we select another feature in the same tree at random with probability proportional to the fraction of data points that used it in that batch.
We then reset the transformation parameters at random, with small variance and centered around the chosen filter's parameters.

\section{Experiments}

\subsection{Learning Representations}
\begin{figure}[ht!]
\begin{center}
    \subfigure[]{
		\quad
		\includegraphics[height=0.37\textwidth,width=0.6\textwidth]{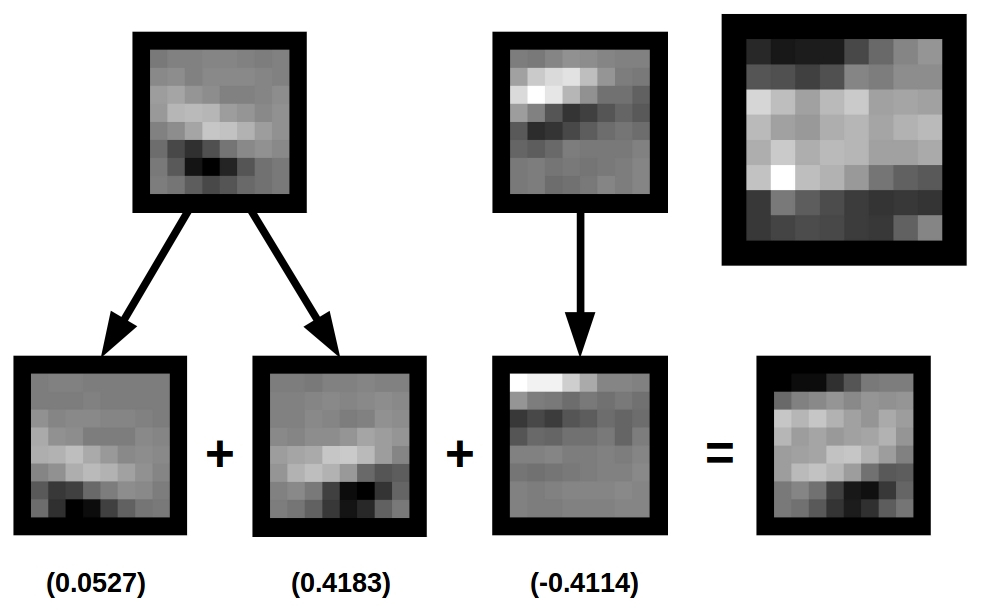}
    }
\end{center}
\caption{Reconstruction example. The root features are transformed and combined with different weights to reconstruct (bottom right) the $8\times 8$ natural
image patch in the top right corner.}
\label{fig:rec_example}
\end{figure}

\begin{figure}[ht!]
\begin{center}
    \subfigure[]{
		\includegraphics[height=0.05\textwidth,width=0.4\textwidth]{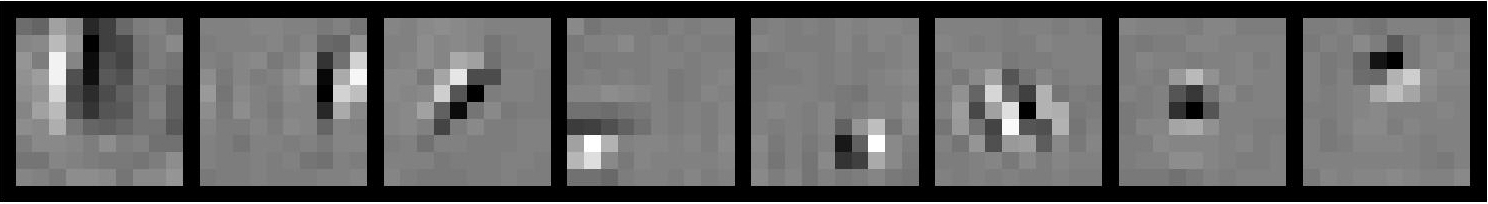}
		\label{res:8by8_roots_example}
    }\\
    \subfigure[]{
		\includegraphics[height=0.4\textwidth,width=0.4\textwidth]{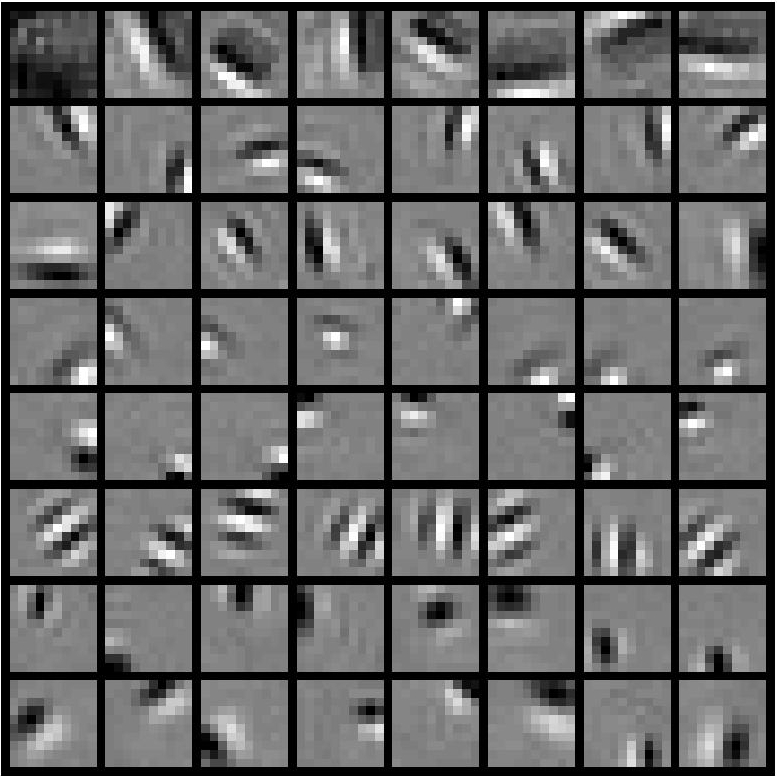}
		\label{res:8by8_leaves_example}
    }
\end{center}
\caption{Learned features for $8$ trees with a branching factor of $8$. (a) Features corresponding to the roots. (b) Features/Leaves: Each row corresponds to leaves/transformations of the same root.}
\label{res:8by8_example}
\end{figure}

We apply transformational sparse coding (TSC) with forests of flat trees to natural image patches. Our approach allows us to learn features resembling
those of traditional sparse coding. Apart from reconstructing the input, the model also extracts transformation parameters from the data.
Figure \ref{fig:rec_example} shows a reconstruction example. Figure \ref{res:8by8_example}
shows the root features learned from $10\times 10$ natural image patches using a forest of size $8$ with branching factor $8$, equipped with the full six-dimensional group.
The forest has a total of $64$ features. Figure \ref{res:8by8_roots_example} shows the features corresponding to the roots. Figure \ref{res:8by8_leaves_example}
shows the corresponding leaves. Each row contains features derived from the same root. More examples of learned features are shown in Figures \ref{fig:features16x16},
\ref{fig:features8x32}, \ref{fig:features4x16} and \ref{fig:features1x64} in Appendix \ref{app:feats}.

\nop{
To determine whether the model learns interesting, large transformations we control for the magnitudes of the group parameters
and their distance from their initialization values. Figure \ref{res:dist_mag} shows these quantities for the $8\times 8$ forest layout.

\begin{figure}[ht!]
\begin{center}
    \subfigure[]{
		\includegraphics[height=0.3\textwidth,width=0.45\textwidth]{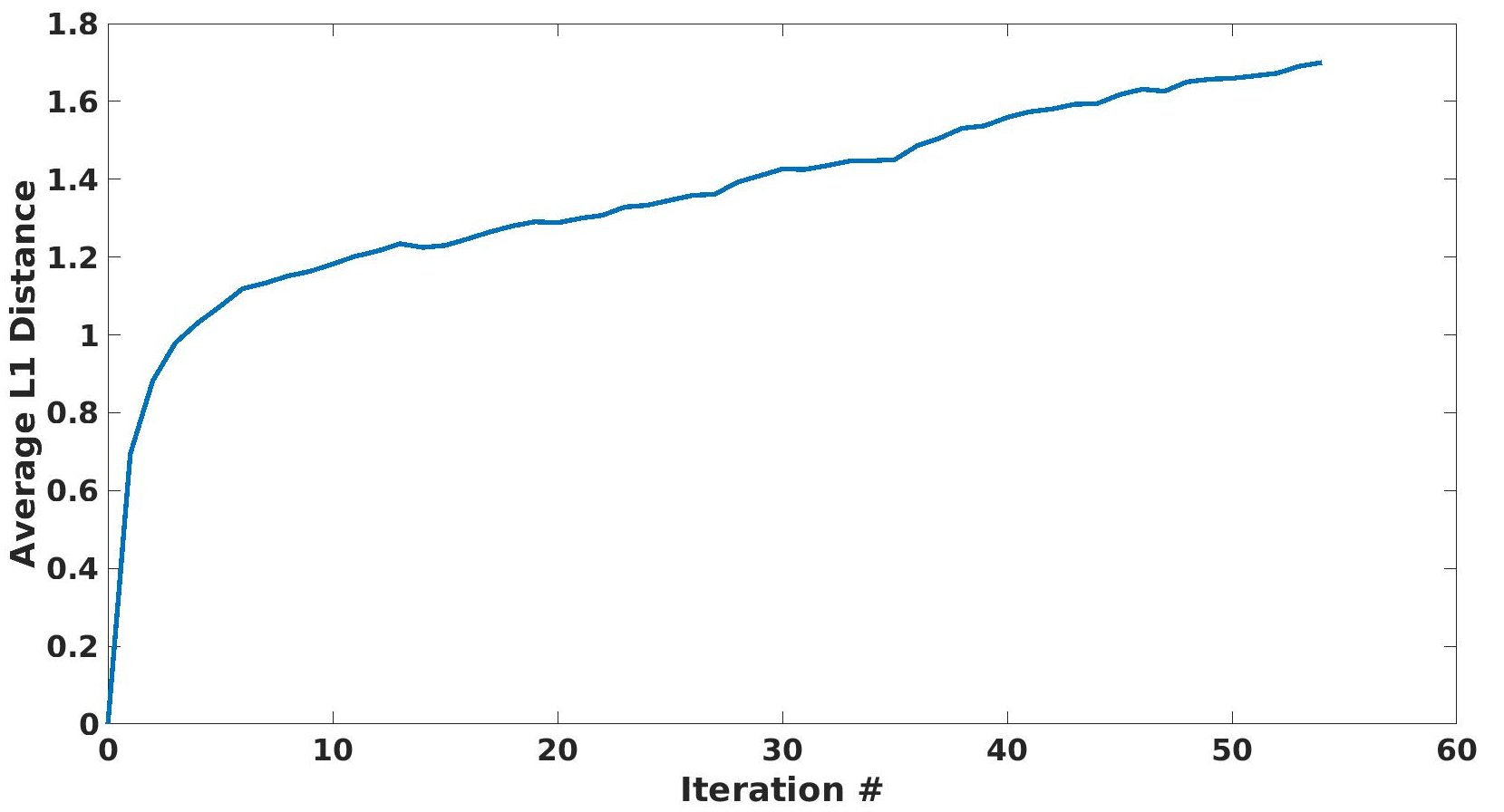}
		\label{res:dist}
    }
    \subfigure[]{
		\includegraphics[height=0.3\textwidth,width=0.45\textwidth]{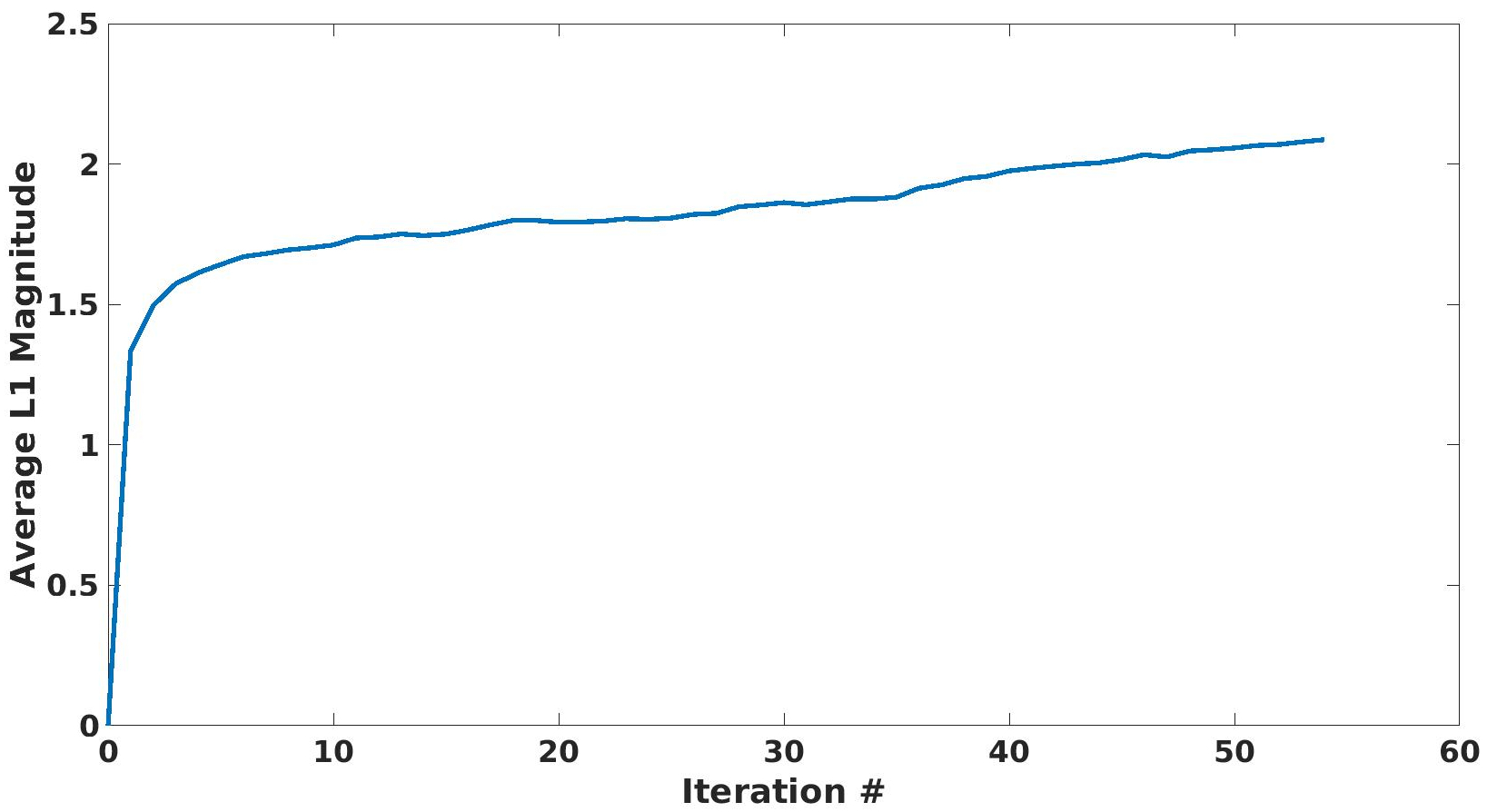}
		\label{res:mag}
    }
\end{center}
\caption{$\ell_1$ distance from initialized values and $\ell_1$ magnitude of the transformation parameters per iteration for the $8\times 8$ forest layout.}
\label{res:dist_mag}
\end{figure}
}

\subsection{Comparison with Sparse Coding}

Even though derivative features have to be explicitly constructed for inference, the degrees of freedom of our model
are significantly lower than that of traditional sparse coding. Specifically:

$$
df_{TSC} = (\text{\# of roots})\times \left(\text{\# of pixels} - 1 + \text{branching factor}\times \text{group dimension}\right) 
$$
whereas:
$$
df_{SC} = \text{\# of features}\times \left(\text{\# of pixels}-1\right)
$$
Note that the group dimension is equal to $3$ for rigid motions and $6$ for general $2$D affine transformations.

We compare transformational sparse coding forests of various layouts and choices for $\lambda_w$ with traditional sparse coding on $10\times 10$ natural image patches.
Some transformations change the feature magnitudes and therefore the sparsity pattern of the weights. To make the comparison clearer, for each choice of layout and penalty coefficient, we run sparse
coding, constraining the feature magnitudes to be equal to the average feature magnitude of our model.
The results are shown in Table \ref{tab:res_tsc_sc}. The reconstruction error of our model is close to that of sparse coding, albeit with slightly less sparse solutions,
even though it has significantly fewer degrees of freedom. Our model extracts pose information in the form of group parameters.

\begin{table}[t!]
	\caption{Comparison of transformational sparse coding (TSC) with sparse coding (SC) for $10\times 10$ natural image patches. We compare the error (MSE) and the degrees of freedom ($df$) over $40000$ data points. ``Sparsity'' is the average number of non-zero weights. $\lambda_w$ is the penalty coefficient for the weights and controls the sparseness of the solution.}
	\label{tab:res_tsc_sc}
	\begin{center}
		\begin{tabular}{c c c c c c c c c c c}
		\multicolumn{1}{c}{} & \multicolumn{4}{c}{\bf TSC} & \multicolumn{4}{c}{\bf SC} &
		\\ \hline 
		$\lambda_w$ & Layout & MSE & Sparsity & $df_{TSC}$ & MSE & Sparsity & $df_{SC}$ & \# of features & $\frac{df_{SC}}{df_{TSC}}$\\\hline \\
		0.4 & $1\times 64$ & 2.13 & 13.3 &  447 & 1.71 & 12.3 & 6336  & 64  & 14.17 \\
		0.5 & $1\times 128$ & 2.28 & 12.1 &  867 & 1.96 & 10.3 & 12672  & 128  & 14.62 \\
		0.4 & $8\times 8$  & 1.89 & 13.3 & 1176 & 1.72 & 12.5 & 6336  & 64  &  5.38 \\
		0.4 & $4\times 16$ & 1.91 & 13.3 &  780 & 1.69 & 12.3 & 6336  & 64  & 8.12 \\
		0.5 & $8\times 8$  & 2.36 & 10.4 & 1176 & 2.15 & 9.9 & 6336  & 64  &  5.38 \\
		0.5 & $4\times 16$  & 2.38 & 11 & 780 & 2.12 & 10.0 & 6336  & 64  & 8.12 \\
		0.4 & $16\times 16$  & 1.66 & 14.3 & 3120 & 1.56 & 13.2 & 25344 & 256 & 8.12  \\
		0.4 & $8\times 32$  & 1.67 & 14.6 & 2328  & 1.56 & 13.2 & 25344 & 256 & 10.88  \\
		\end{tabular}
	\end{center}
\end{table}

\nop{
\subsection{Fine-Tuning}

Feature transformations learned by our model are common across all images. It is expected that fitting each feature on each image
independently would yield a better reconstruction. However, as we discussed in Section \ref{model}, this makes learning larger
transformations hard. A compromise between the two would involve ``fine-tuning'' at the leaves. We can model the images as a combination
of common features/leaves that are further transformed slightly for better reconstruction. The image is then modeled as:

$$
I_i \simeq  \sum_{v=1}^{V} \sum_{b\sim ch(v)}w_{ib}T(x_i+x_{v\rightarrow b})F_v
$$
Since optimizing the full matrix exponential is unwieldy, if we constraint $x_i$ to be small, we can approximate the transformation
above by the transformation associated with the leaf, chained with a first-order Taylor approximation of the fine-tuning transformation:
$$
T(x_i+x_{v\rightarrow b}) \simeq \left(\mathbb{I} + \sum_{j=1}^{6} x_{ij}G_j\right) T(x_{v\rightarrow b})
$$
We can then optimize $x_i$ to obtain $\hat{x}_i$. To recover a single set of parameters for each feature transformation we can
approximate again:
$$
\left(\mathbb{I} + \sum_{j=1}^{6} \hat{x}_{ij}G_j\right)T(x_{v\rightarrow b}) \simeq T(\hat{x}_i+x_{v\rightarrow b})
$$
For the $8\times 8$ and $4\times 16$ instances, this improves the error by $1$ and $1$ respectively.
}

\subsection{Deeper Trees}
\label{sec:deeper_trees}
We can define deeper trees by associating a set of transformation parameters with each branch. These correspond to additive contributions
to the complete transformation that yields the leaf when applied to the root:
$$
I_i \simeq \sum_{v=1}^{V} \sum_{b\sim ch(v)}w_{ib}T(x_P)F_v
$$
where $x_P = \sum_{e\in \text{path}(b,v)} x_e$. Optimizing deeper trees is more demanding due to the increased number of parameters. Their advantage
is that they lend structure to model. The parameters corresponding to the subtree of an internal node tend to explore the parameter subspace close
to the transformation defined by that internal node. In tasks where it is disadvantageous to marginalize completely over transformations, equivariant
representations corresponding to intermediate tree layers can be used. An example of such structure is shown in Figure \ref{fig:flat_vs_deep} in Appendix
\ref{app:deep}.

\section{Related Work}
\cite{DBLP:journals/corr/abs-1001-1027} present a model for fitting Lie groups to video data. Their approach
only works for estimating a global transformation between consecutive video frames. They only support transformations of a single kind (ie only rotations). Different
such single-parameter transformations have to be chained together to produce the global one. The corresponding transformation parameters also have to be inferred
and stored in memory and cannot be directly converted to parameters of a single transformation. \cite{4906239} present an approach to optimally estimating transformations between pairs of images. They
support rigid motions and isotropic scaling. \cite{NIPS2009_3791} focus on learning the group operators and transformation parameters from pairs of images, but do not learn features from data.
Our model supports all six transformations and learns object parts and their individual
transformations. In contrast with those approaches, our model learns object parts jointly with their transformations within the same image. Our model utilizes the full, six-dimensional,
general affine Lie group and captures the pose of each object part in the form of a single set of six transformation parameters.

\cite{Grimes:2005:BSC:1119610.1119615} propose a bilinear model that combines sparse coding with transformations. The model accounts for global transformations that apply to the entire image
region. Our model accounts for individual transformations of image parts. \cite{doi:rao_1998_what_where} propose a model that captures small image transformations with Lie groups using a first-order Taylor approximation. Our model estimates larger transformations of image parts using the full exponential model. \cite{Rao:1999:LLG:340534.340807} and \cite{citeulike:2714601} use a first-order Taylor
approximation to learn the group operators and the transformation parameters for small transformations.

The work closest to ours is that of \cite{Hinton:2011:TA:2029556.2029562} on capsules. A capsule learns to recognize its template (feature) over a wide range of poses. The pose is computed
by a neural network (encoder). The decoder, resembling a computer graphics engine combines the capsule templates in different poses to reconstruct the image. Each transformational sparse coding
tree can be thought of as capsule. The template corresponds to the root. The tree learns to ``recognize'' transformed versions of that template. Our work arrives at the concept of a capsule
from a sparse coding perspective. A major difference is that our approach allows us to reuse each feature multiple times in different, transformed versions for each data point.

\cite{Gens:2014:DSN:2969033.2969110} propose a convolutional network that captures symmetries in the data by modeling symmetry groups. Experiments with rigid motions or various affine transformations
show reduced sample complexity. \cite{DBLP:journals/corr/CohenW16} propose a convolutional network that can handle translations, reflections and rotations of $90$ degrees. \cite{DBLP:journals/corr/DielemanFK16}
propose a network that handles translations and rotations. All the above are supervised learning models and apart from the first can handle a limited set of transformations. Our model is completely
unsupervised, extends sparse coding and can handle all transformations given by the first order differential equation:
$$
\frac{\partial I(\theta)}{\partial \theta} = A I(\theta)
$$
as described by \cite{Rao:1999:LLG:340534.340807}.
\section{Conclusion}
In this paper, we proposed a sparse coding based model that learns object features jointly with their transformations, from data.
Naively extending sparse coding for data-point specific transformations makes inference intractable. We introduce a new technique that circumvents
this issue by using a tree structure that represents common transformations in data. We show that our approach can learn interesting
features from natural image patches with performance comparable to that of traditional sparse coding.

Investigating the properties of deeper trees, learning the tree structure dynamically from the data and extending our model into
a hierarchy are subjects of ongoing research.

\bibliography{TSC_arxiv}
\bibliographystyle{iclr2017_conference}

\renewcommand{\thesubsection}{\Alph{subsection}}

\begin{appendices}

\section*{Appendix}
\subsection{Generator Effects}
\label{app:Gs}
\begin{figure}[ht!]
\label{app:feats}
\begin{center}
    \subfigure[]{
		\includegraphics[height=0.2\textwidth,width=0.2\textwidth]{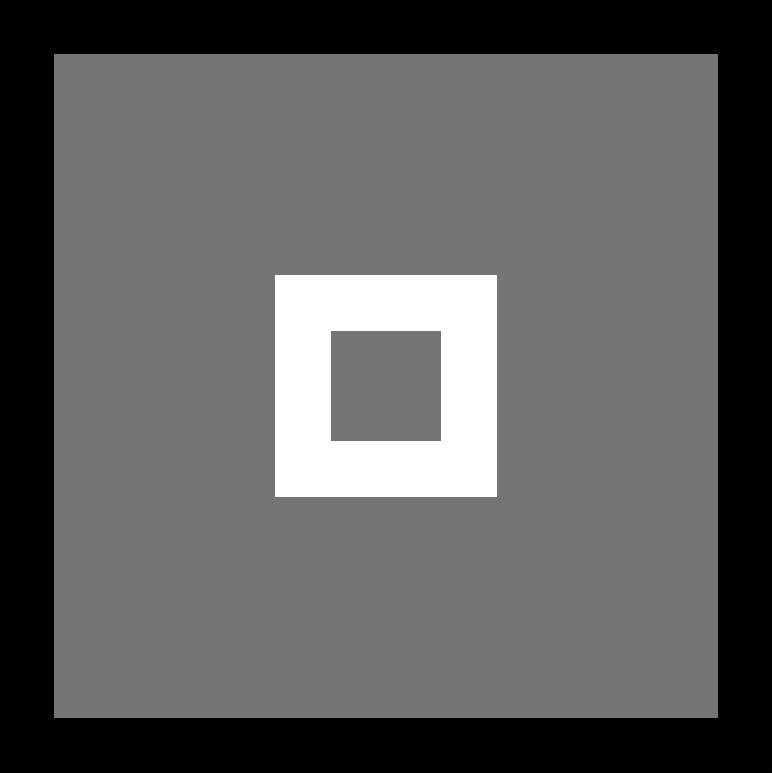}
		\label{fig:G_temp}
    }\\
    \subfigure[]{
		\includegraphics[height=0.06\textwidth,width=0.8\textwidth]{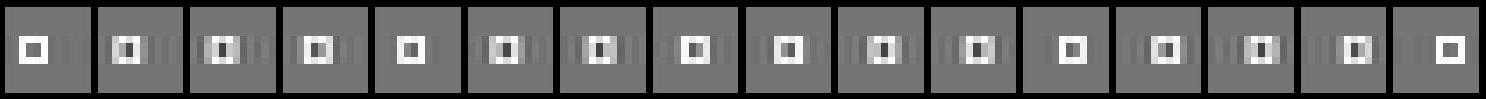}
		\label{fig:G_tx}
    }\\
    \subfigure[]{
		\includegraphics[height=0.06\textwidth,width=0.8\textwidth]{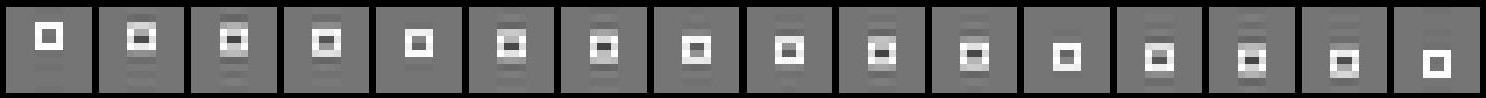}
		\label{fig:G_ty}
    }\\
    \subfigure[]{
		\includegraphics[height=0.06\textwidth,width=0.8\textwidth]{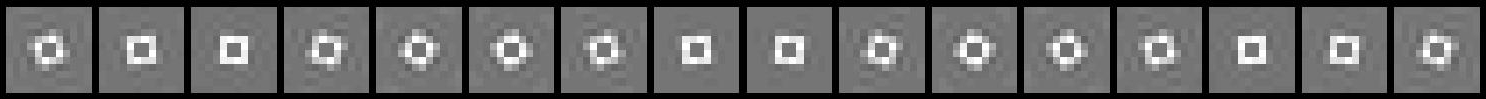}
		\label{fig:G_rot}
    }\\
    \subfigure[]{
		\includegraphics[height=0.06\textwidth,width=0.8\textwidth]{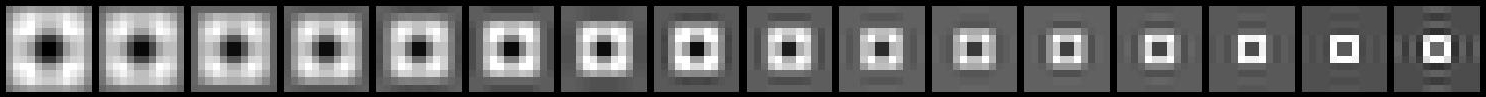}
		\label{fig:G_sc}
    }\\
    \subfigure[]{
		\includegraphics[height=0.06\textwidth,width=0.8\textwidth]{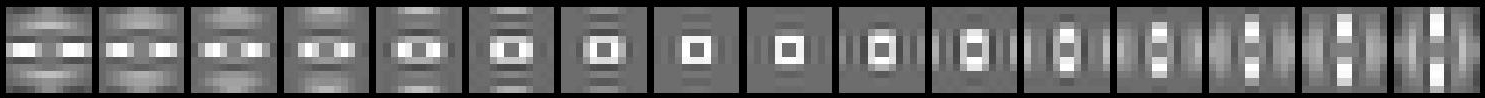}
		\label{fig:G_h1}
    }\\
    \subfigure[]{
		\includegraphics[height=0.06\textwidth,width=0.8\textwidth]{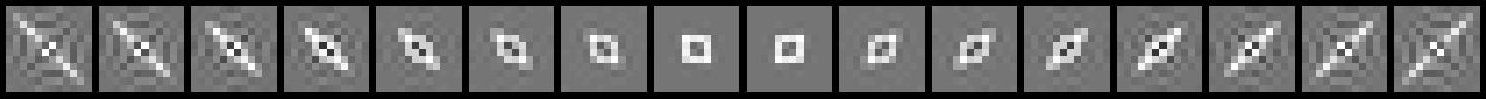}
		\label{fig:G_h2}
    }
\end{center}
\caption{Effects of each individual transformation on the template (a): (b) horizontal translation, (c) vertical translation, (d) rotation, (e) scaling, (f) parallel hyperbolic deformation along the $X/Y$
axis, (g) hyperbolic deformation along the diagonals. To compute the generators, we used the $\sinc$ interpolation function.}
\label{fig:G_effects}
\end{figure}

Figure \ref{fig:G_effects} presents the effects of each individual transformation of the six that are supported by our model.
The template is a square \ref{fig:G_temp}.

\subsection{Deeper trees and structure}
\label{app:deep}

\begin{figure}[ht!]
\begin{center}
    \subfigure[]{
		\includegraphics[height=0.4\textwidth,width=0.8\textwidth]{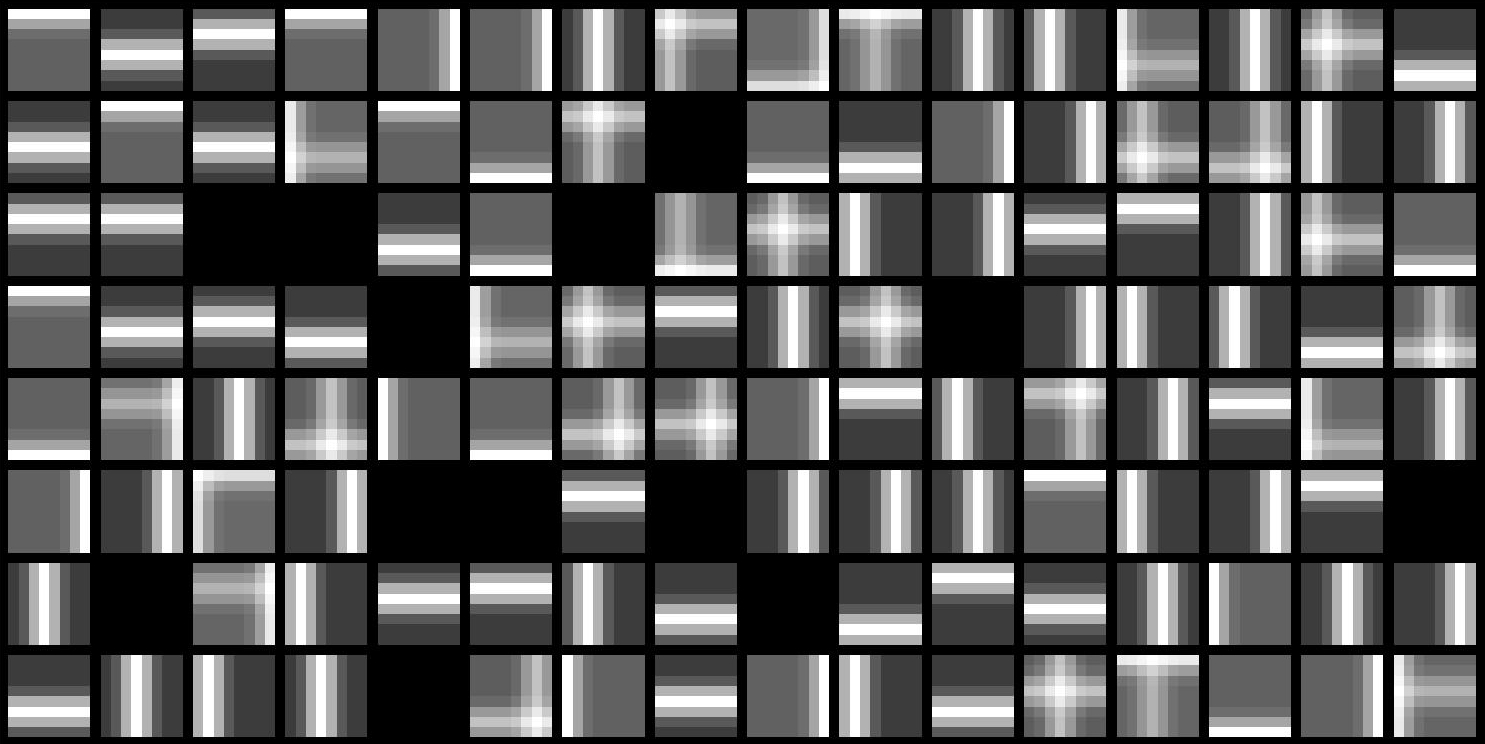}
		\label{fig:vs_patches}
    }\\
    \subfigure[]{
		\includegraphics[height=0.06\textwidth,width=0.8\textwidth]{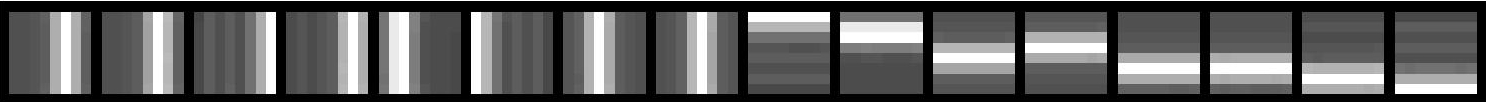}
		\label{fig:vs_flat}
    }\\
    \subfigure[]{
		\includegraphics[height=0.06\textwidth,width=0.8\textwidth]{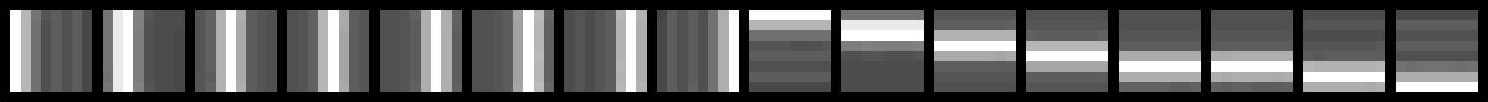}
		\label{fig:vs_deep}
    }
\end{center}
\caption{Features learned for the double-line example: (a) Input, (b) features learned by a forest of two flat trees of size eight, (c) features learned by two binary trees of the same size.
For (c) the leaves have been reordered with subtree permutations to reveal the order. Each subtree learns features corresponding to an area of the input.}
\label{fig:flat_vs_deep}
\end{figure}

Figure \ref{fig:flat_vs_deep} presents an example of structure learned by deeper trees. This example consists of vertical and horizontal lines.
Each image patch is either blank, contains one vertical or one horizontal line or both. A patch is blank with probability $\frac{1}{9}$,
contains exactly one line with probability $\frac{2}{3}$ or two lines with probability $\frac{2}{9}$. Each line is then generated at one
of eight positions at random. Fitting two binary trees results in some continuity in the features, whereas flat trees
provide no such structure.

\subsection{Learned Features}
\label{app:feats}
\begin{figure}[ht!]
\begin{center}
    \subfigure[]{
		\includegraphics[height=0.4\textwidth,width=0.4\textwidth]{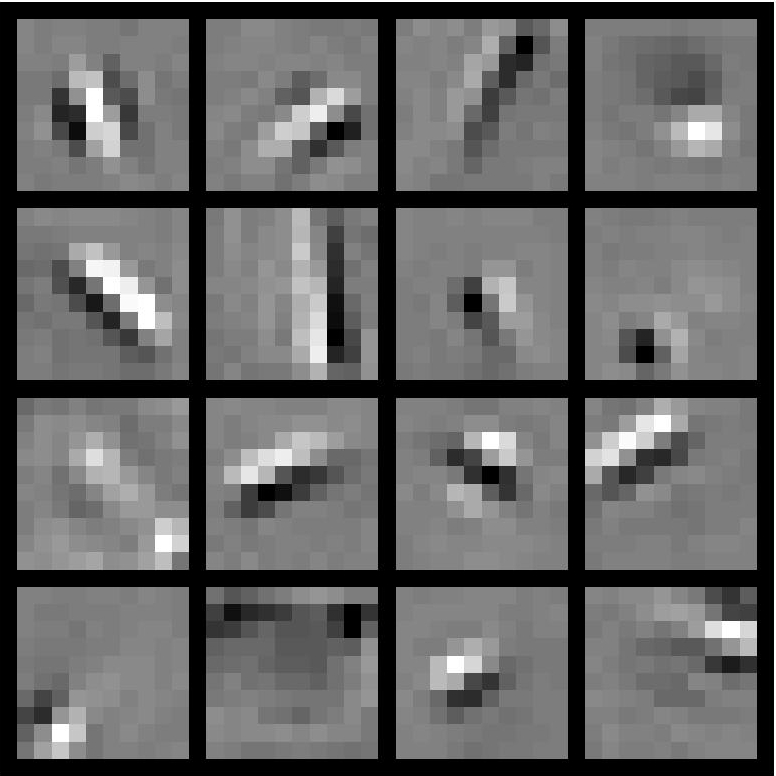}
		\label{fig:features16x16_roots}
    }\\
    \subfigure[]{
		\includegraphics[height=0.8\textwidth,width=0.8\textwidth]{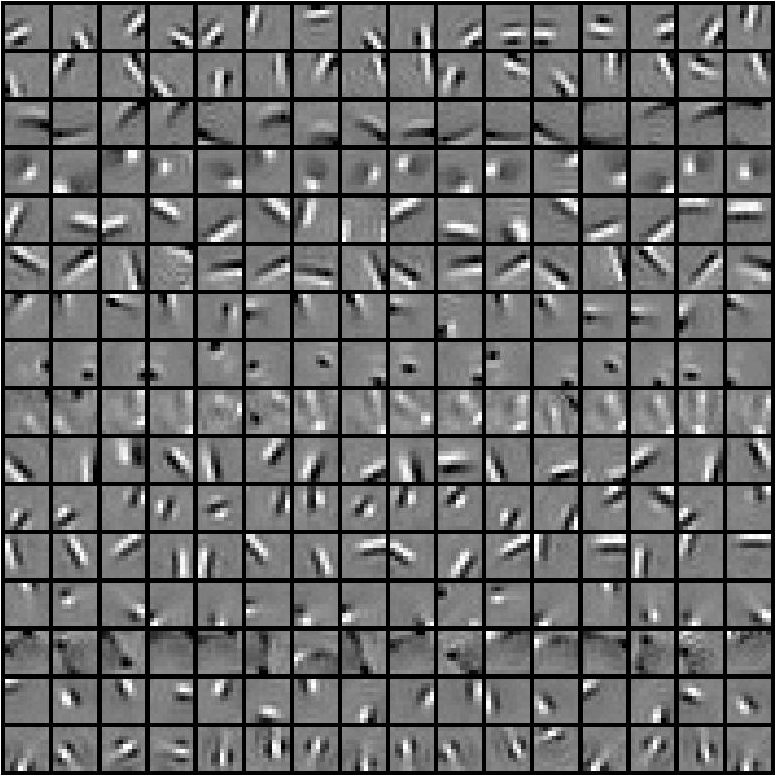}
		\label{fig:features16x16_leaves}
    }
\end{center}
\caption{Learned features for $16$ trees with branching factor $16$. Each row corresponds to leaves/transformations of the same root.}
\label{fig:features16x16}
\end{figure}
\begin{figure}[ht!]
\begin{center}
    \subfigure[]{
		\includegraphics[height=0.1\textwidth,width=0.8\textwidth]{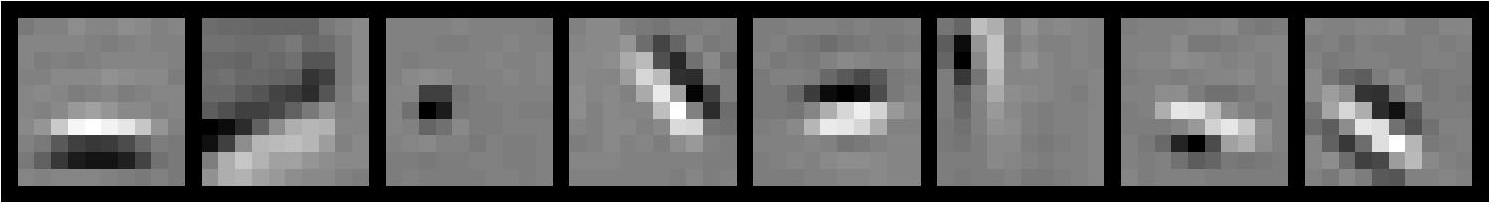}
		\label{fig:features8x32_roots}
    }\\
    \subfigure[]{
		\includegraphics[height=0.25\textwidth,width=1\textwidth]{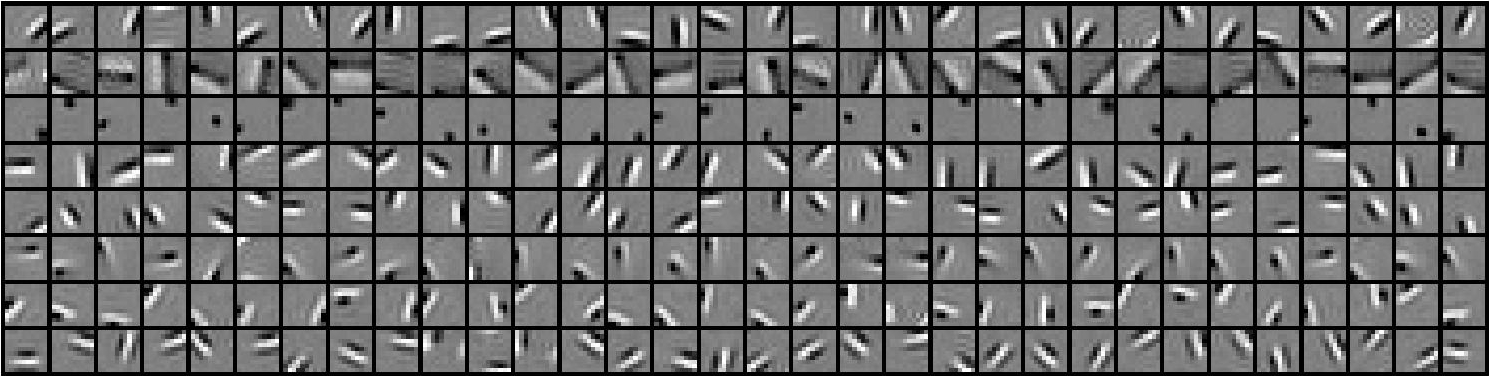}
		\label{fig:features8x32_roots}
    }
\end{center}
\caption{Learned features for $8$ trees with branching factor $32$. Each row corresponds to leaves/transformations of the same root.}
\label{fig:features8x32}
\end{figure}
\begin{figure}[ht!]
\begin{center}
    \subfigure[]{
		\includegraphics[height=0.1\textwidth,width=0.4\textwidth]{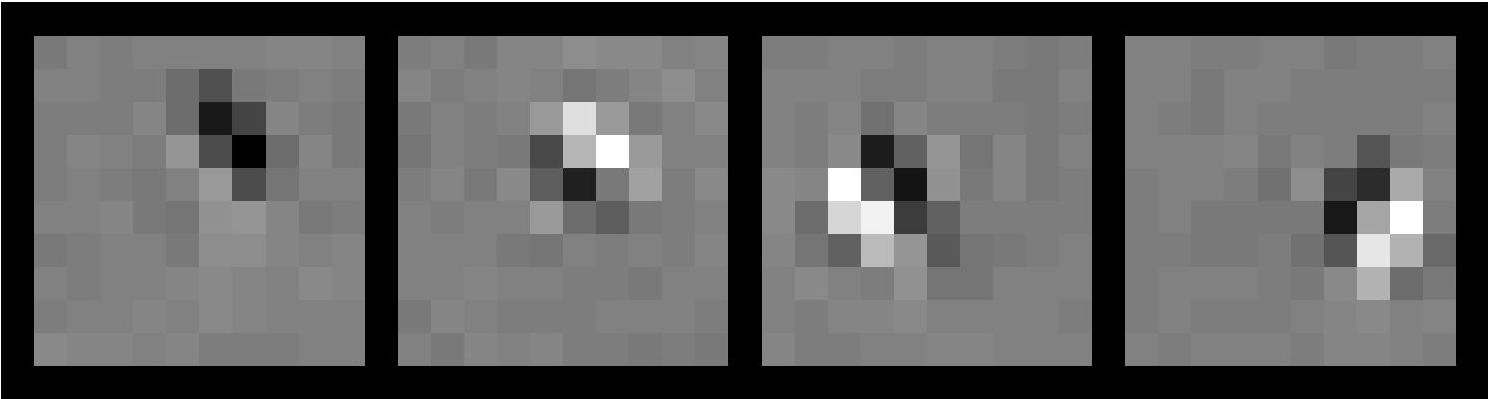}
		\label{fig:features4x16_roots}
    }\\
    \subfigure[]{
		\includegraphics[height=0.25\textwidth,width=1\textwidth]{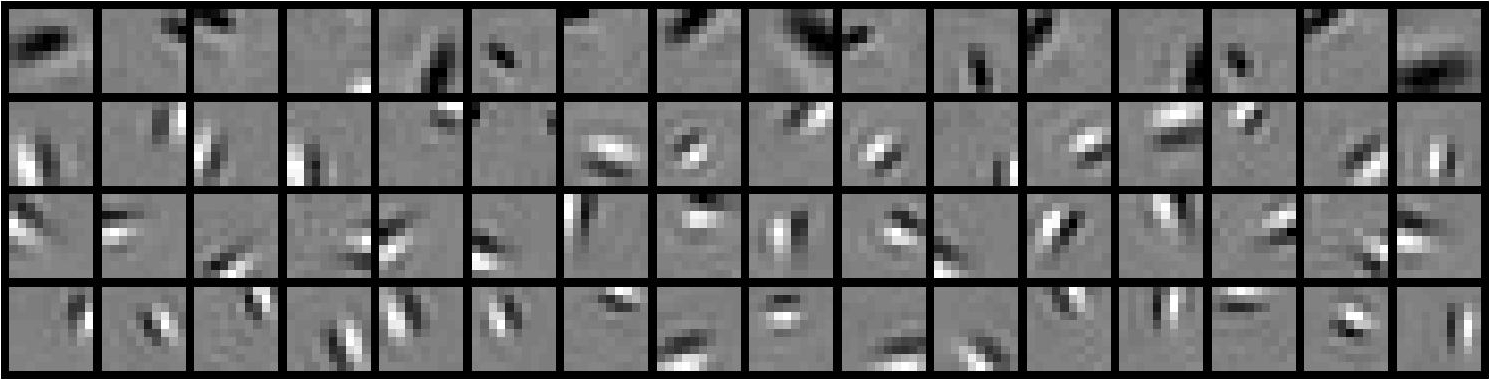}
		\label{fig:features4x16_roots}
    }
\end{center}
\caption{Learned features for $4$ trees with branching factor $16$. Each row corresponds to leaves/transformations of the same root.}
\label{fig:features4x16}
\end{figure}
\begin{figure}[ht!]
\begin{center}
    \subfigure[]{
		\includegraphics[height=0.1\textwidth,width=0.1\textwidth]{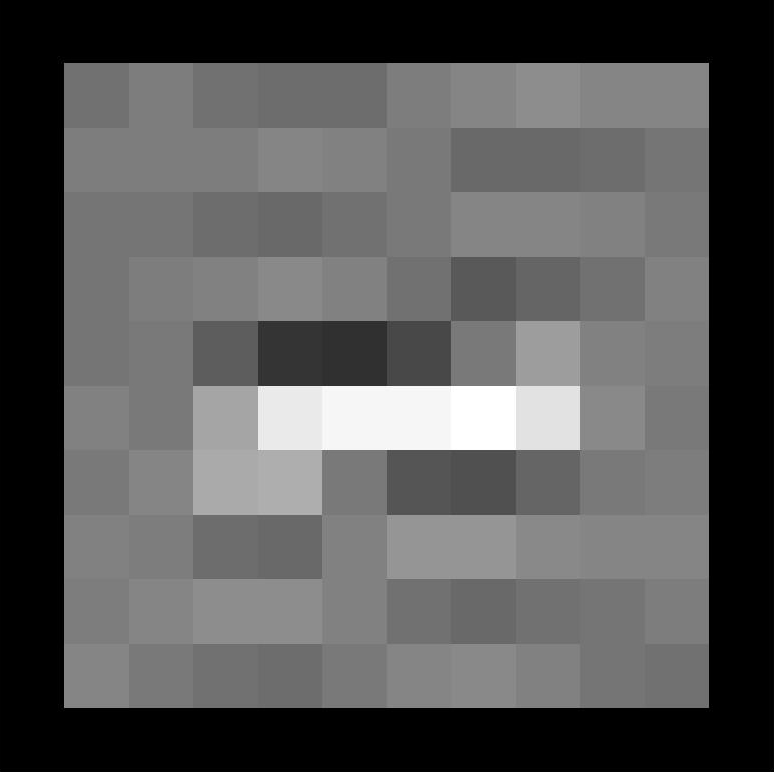}
		\label{fig:features1x64_roots}
    }\\
    \subfigure[]{
		\includegraphics[height=0.4\textwidth,width=0.4\textwidth]{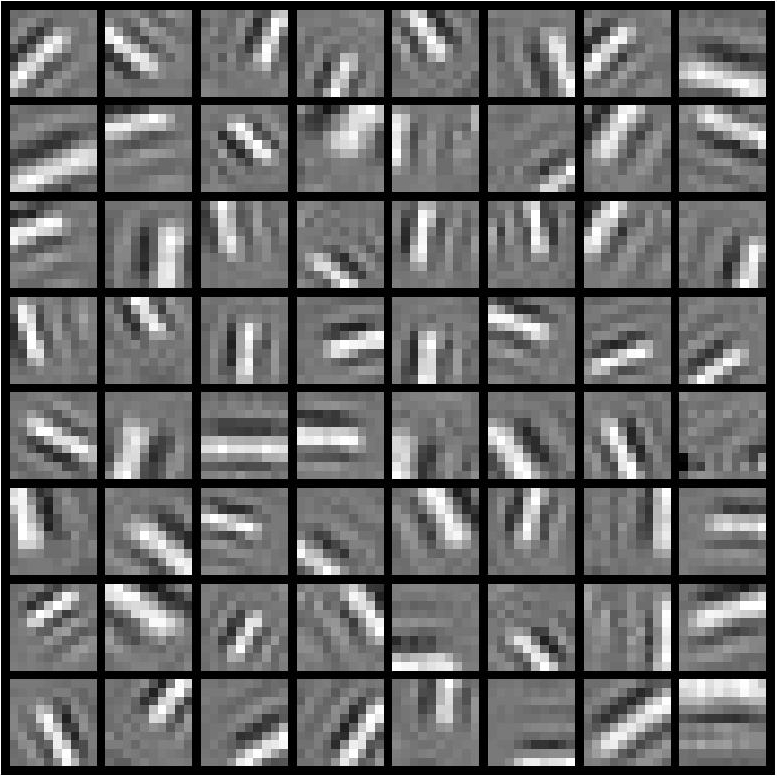}
		\label{fig:features1x64_roots}
    }
\end{center}
\caption{Learned features for $1$ tree with branching factor $64$. All features are transformations of the same root.}
\label{fig:features1x64}
\end{figure}
\end{appendices}
\end{document}